\pdfoutput=1
\documentclass[11pt]{article}

\usepackage{ACL2023}

\usepackage{times}
\usepackage{latexsym}
\usepackage[T1]{fontenc}
\usepackage[utf8]{inputenc}
\usepackage{microtype}
\usepackage{inconsolata}
\usepackage{multirow}
\usepackage{amsfonts}
\usepackage{xcolor}
\usepackage{amssymb}
\usepackage{pifont}
\usepackage{colortbl}
\usepackage{amsmath}
\usepackage{pifont}
\usepackage{graphicx}
\usepackage{booktabs}
\usepackage{amsthm}
\usepackage[ruled,vlined,linesnumbered,noresetcount]{algorithm2e}
\SetKwRepeat{Do}{do}{while}
\SetKw{And}{and}
\SetKw{In}{in}

\usepackage{color}
\usepackage{enumitem}

\theoremstyle{plain}
\newtheorem{theorem}{Theorem}[section]

\theoremstyle{definition}
\newtheorem{notation}[theorem]{Notation}
\newtheorem{definition}[theorem]{Definition}
\theoremstyle{remark}

\title{Rethinking Cross-Subject Data Splitting for Brain-to-Text Decoding}

\author{Congchi Yin\textsuperscript{\rm 1,2}$^{\ast}$, Qian Yu\textsuperscript{\rm 3}, Zhiwei Fang\textsuperscript{\rm 3}, 
{\bf Changping Peng\textsuperscript{\rm 3},
Piji Li\textsuperscript{\rm 1,2}$^{\dag}$}\\
  \textsuperscript{\rm 1} 
  College of  Artificial Intelligence, Nanjing University of Aeronautics and Astronautics \\
  \textsuperscript{\rm 2} The Key Laboratory of Brain-Machine Intelligence Technology, Ministry of Education \\
  \textsuperscript{\rm 3} JD.com \\
  \texttt{\{congchiyin,pjli\}@nuaa.edu.cn}}
\begin{document}
\renewcommand{\arraystretch}{0.9}

\maketitle
\renewcommand{\thefootnote}{\fnsymbol{footnote}}
\footnotetext[1]{Work done during internship in JD.com.}
\footnotetext[2]{Corresponding author.}
\renewcommand{\thefootnote}{\arabic{footnote}}


\begin{abstract}
Recent major milestones have successfully reconstructed natural language from non-invasive brain signals (e.g. functional Magnetic Resonance Imaging (fMRI) and Electroencephalogram (EEG)) across subjects.
However, we find current dataset splitting strategies for cross-subject brain-to-text decoding are wrong. Specifically, we first demonstrate that all current splitting methods suffer from data leakage problem, which refers to the leakage of validation and test data into training set, resulting in significant overfitting and overestimation of decoding models.
In this study, we develop a right cross-subject data splitting criterion without data leakage for decoding fMRI and EEG signal to text. 
Some SOTA brain-to-text decoding models are re-evaluated correctly with the proposed criterion for further research. 
\end{abstract}

\section{Introduction}

Brain-to-text decoding aims to recover natural language from brain signals stimulated by corresponding speech. Recent studies \citep{makin2020machine,DBLP:conf/aaai/WangJ22,DBLP:conf/acl/XiZWL0023,tang2023semantic,duan2024dewave} have successfully decoded non-invasive brain signals (e.g. fMRI, EEG) to text by applying deep neural networks. Most of these works perform within-subject data splitting for training and evaluating decoding models. This subject-specific splitting method causes two main problems. First, it only uses data from one subject of the whole dataset for training and testing. Since brain signal collection is costly and time-consuming, such splitting method results in a significant waste of data resources. Second, it leads to poor model generalization. As every brain has unique functional and anatomical structures, subject-specific models may exhibit considerable variability across individuals and fail to generalize to other subjects \citep{liu2024see}. Moreover, decoding models trained from scratch on limited data are prone to facing the overfitting problem.

Human brain responds similarly to the same stimuli, despite the individual discrepancy \citep{hasson2004intersubject,pereira2018toward}. Therefore, some studies \citep{DBLP:conf/aaai/WangJ22,DBLP:conf/acl/XiZWL0023,duan2024dewave} begin to shed light on cross-subject brain-to-text decoding, which performs data splitting based on all the subjects, trains and evaluates decoding model once. Cross-subject data splitting effectively compensates for the shortcomings of subject-specific splitting, and has been widely applied in brain-to-image decoding \citep{wang2024mindbridge,liu2024see}. However, 
unlike datasets for brain-to-image decoding \citep{allen2022massive,chang2019bold5000} where subjects are guided to see different and unrepeated pictures, different subjects will be stimulated by the same story in common naturalistic language comprehension dataset, which challenges cross-subject data splitting.

Based on our observations, current cross-subject data splitting methods for brain-to-text decoding are wrong because data for validation and test leaks into the training set, rendering the evaluation of the decoding process meaningless. Specifically, we find two types of data leakage: \textit{brain signal leakage} and \textit{text stimuli leakage}. Brain signal leakage refers to test subject’s brain signal appears in training set. Text stimuli leakage refers to text in test set appears in the training set.
Modern brain-to-text decoding models follow an encoder-decoder manner. We pick two representative models: EEG2Text \citep{DBLP:conf/aaai/WangJ22} and UniCoRN \citep{DBLP:conf/acl/XiZWL0023} to reveal data leakage and its damage. 
Experiments support that data leakage affects model training on both encoder and decoder side.
For the encoder, it will become overfitting and fail to well represent brain signals if brain signal leakage exists. For the decoder, the situation gets worse if text stimuli leakage happens. Any data leakage would cause the auto-regressive decoder to memorize previously seen paragraphs during training stage, resulting in poor generalization to unseen text. 

To avoid data leakage and fairly evaluate the performance of cross-subject brain-to-text decoding models, we propose a right data splitting method. We focus on fMRI and EEG signals in this study, although the proposed criterion could be applied to any datasets satisfying the prescribed format. In the proposed method, we follow two basic rules: (1) Brain signals collected from specific subject in validation set and test set will not appear in training set, which means the trained encoder cannot get access to any brain information belonging to subjects in test set. (2) Text stimuli in validation set and test set will not appear in training set. The decoder learns to reconstruct language with brain signals instead of memorizing seen text.

Our contributions can be summarized as follows:
\begin{itemize}[topsep=0pt]
\setlength\itemsep{-0.5em}
    \item To the best of our knowledge, we are the first to identify the issue of data leakage in current cross-subject data splitting methods for brain-to-text decoding.
    \item We define the splitting criterion for cross-subject brain-to-text decoding, and propose a right dataset splitting method.
    \item Some SOTA brain-to-text decoding models are re-evaluated using the proposed cross-subject data splitting method to ensure a fair assessment of their performance.
\end{itemize}

\section{Preliminary} \label{formu}
\subsection{Dataset Description} \label{dataset}
A naturalistic language comprehension dataset $\mathcal{D}$ contains brain signals of $N$ subjects when they passively listen to $K$ spoken stories. Suppose that not all subjects are stimulated by all stories, and different subjects may hear the same story. 


Formally, $S_1, S_2, \ldots, S_N$ denotes to the $N$ subjects and $M_1, M_2, \ldots, M_K$ denotes to the $K$ stories in dataset. The $k$-th story $M_k$ consists of $l_{k}$ text segments $T_{k1}, T_{k2}, \ldots, T_{kl_k}$. If the $i$-th subject $S_i$ hears the $j$-th text segment $T_{kj}$, then his brain signal is denoted as $F_{ijk}$.

\begin{figure}
\centering
\includegraphics[width=0.49\textwidth]{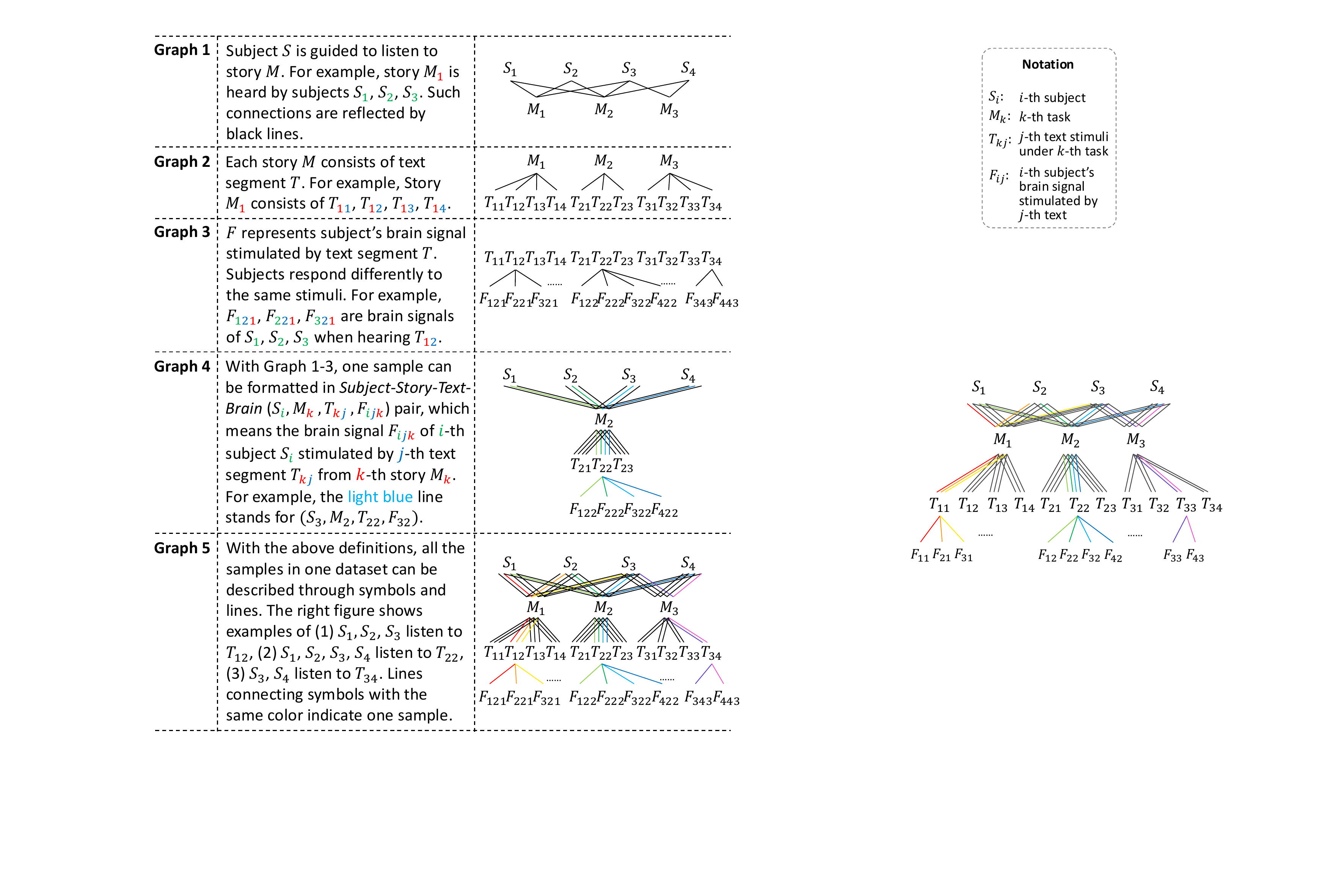}
\caption{Illustration of how to build graph to describe dataset step by step.}
\label{f1}
\vspace{-3mm}
\end{figure}

\subsection{Use Graph to Describe Dataset} \label{graphdef}
We use \textit{multigraph} and $k$\textit{-partite graph} (detailed in \ref{multi} \& \ref{k}) to describe the intricate structure of naturalistic language comprehension dataset. 
\begin{definition}
    A naturalistic language comprehension dataset $\mathcal{D}$ can be represented via a directed $4$\textit{-partite multigraph} $\mathcal{G}_\mathcal{D}$.
\end{definition}

How to build the directed $4$\textit{-partite multigraph} $\mathcal{G}_\mathcal{D}$ step by step is shown in Figure \ref{f1}. Graph 1 is a $2$\textit{-partite} graph indicating subject $S_i$ listening to story $M_k$. Subject $S_i$ and story $M_k$ are viewed as vertices, and edges connecting them indicate certain type of relationship (e.g. $S_i$ ``listen to'' $M_k$ in this case). Graph 2 illustrates that story $M_k$ consists of text segments $T_{kj}$. Graph 3 shows the brain signals $F_{ijk}$ of subject $S_i$ stimulated by text segment $T_{kj}$.
Graph 4 is an example of combining the three $2$\textit{-partite} graphs Graph 1-3: $F_{122},F_{222},F_{322},F_{422}$ are brain signals of $S_1,S_2,S_3,S_4$ stimulated by text segment $T_{22}$ from story $M_2$. In this example, four edges between $M_{2}$ and $T_{22}$ correspond to the different responses of four subjects to the same text segment. There are three edges between $S_2$ and $M_{2}$ because $M_{2}$ contains three text segments. 
Edges of the same color indicate one sample in dataset.
Graph 5 shows the complete directed $4$\textit{-partite multigraph} $\mathcal{G}_\mathcal{D}$ for representing whole dataset. Every sample in dataset can be represented through ordered subject-story-text-brain ($S_i,M_k,T_{kj},F_{ijk}$) pair.
We introduce the formal notation of $\mathcal{G}_\mathcal{D}$:

\begin{figure*}
\centering
\includegraphics[width=\linewidth]{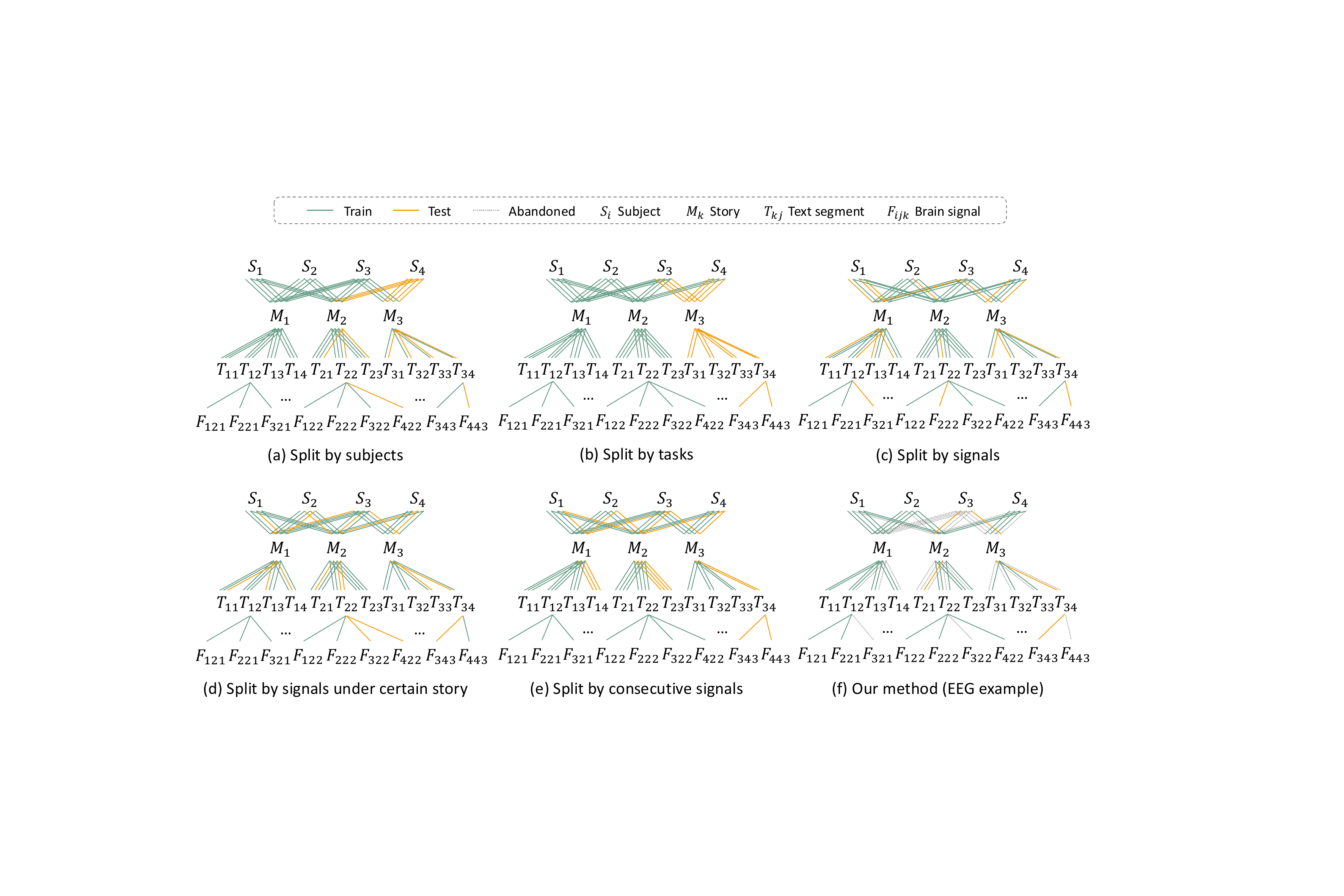}
\caption{Different splitting methods for cross-subject brain-to-text decoding. (Color printing is preferred.)}
\label{f2}
\vspace{-1mm}
\end{figure*}

\begin{notation}
    $\mathcal{G}_\mathcal{D}=(\mathcal{V}, \mathcal{E}, f)$, where $\mathcal{V}=\mathcal{S} \cup \mathcal{M} \cup\mathcal{T} \cup\mathcal{F}$, $\mathcal{S}=\{S_i\}_{i=1}^N$, $\mathcal{M}=\{M_k\}_{k=1}^K$, $\mathcal{T}=\{T_{kj}\}_{k,j=1}^{K,l_k}$, $\mathcal{F}=\{F_{ijk}\}_{i,j,k=1}^{N,l_k,K}$ denote subject set, story set, text segment set, and brain signal set. $f: \mathcal{E} \rightarrow \mathcal{V}\otimes \mathcal{V}$ is an incidence function that maps each edge to a pair of vertices.
\end{notation}








\subsection{Brain-to-Text Decoding Task}
\label{2.3}
The brain-to-text decoding task seeks to build a decoding model that reconstructs natural language text from brain signals, with the goal of accurately decoding what the subject hears. 
Take fMRI and EEG signal for example. fMRI captures brain responses at second level whereas EEG samples brain activity at millisecond level. So the pre-processing for fMRI and EEG input varies. Previous practice in fMRI-to-text decoding \citep{tang2023semantic,DBLP:conf/acl/XiZWL0023} concatenated $L$ future fMRI frames with text segments to form one sample:
\begin{align}
    T_{k,j}^* &= concat(T_{k,j}, T_{k,j+1}, \ldots, T_{k,j+L}), \\
    F_{i,j,k}^* &= concat(F_{i,j,k}, F_{i,j+1,k},\ldots,F_{i,j+L,k}).
\end{align}
In this case, one $(S_i,M_k,T_{kj},F_{ijk})$ pair in graph $\mathcal{G}_\mathcal{D}$ only represents the start point of one sample, while $(S_i,M_k,T_{kj}^*,F_{ijk}^*)$ indicates the whole sample. In EEG-to-text decoding, previous methods sampled continuous EEG signal $F_{ijk}$ that corresponds to text $T_{kj}$. So one $(S_i,M_k,T_{kj},F_{ijk})$ pair is viewed as one sample in our definition.

\section{Methodology} \label{method}

\subsection{Cross-Subject Data Splitting Criterion}
Consistent with cross-subject brain-to-image decoding \citep{wang2024mindbridge,liu2024see}, the dataset splitting should obey two basic principles: (1) If brain signal $F_{ijk}$ appears in test set, then any brain signal $F_{i*k}$ belonging to this subject $i$ should not appear in training set. (2) If text segment $T_{kj}$ appears in test set, then it should not appear in training set. Consistent with the definitions in Section \ref{formu}, graph $\mathcal{G}_\mathcal{D}$ is applied to describe data splitting. Since the validation samples are split in the same manner as the test samples, we focus solely on the test set. Therefore, we have $\mathcal{G}_\mathcal{D}=\mathcal{G}_{train}\cup\mathcal{G}_{test}$.



\subsection{Analysis of Current Splitting Methods}
Edges with different colors are used to represent their classification as either part of the training set or the test set. As shown in Figure \ref{f2}, $(S_i,M_k,T_{kj},F_{ijk})$ pairs with green edges indicate training samples, and those with orange edges are test samples.
Current cross-subject data splitting methods \citep{DBLP:conf/aaai/WangJ22,DBLP:conf/acl/XiZWL0023} can be summarized as five categories:
\begin{itemize}[topsep=0pt]
\setlength\itemsep{-0.5em}
    \item Method (a): Split subjects $\mathcal{S}$ randomly with given ratio.
    \begin{equation}
    \begin{aligned} \label{eq1}
         &\mathcal{G}_{train} = \{(S_i,M_k,T_{kj},F_{ijk} )| \\
          & \forall (S_i^\prime,M_k^\prime,T_{kj}^\prime,F_{ijk}^\prime ) \in \mathcal{G}_{test}, S_i \neq S_{i}^\prime \}
    \end{aligned}
    \end{equation}
    \item Method (b): Split stories $\mathcal{M}$ randomly with given ratio.
    \begin{equation}
    \begin{aligned} \label{eq2}
         &\mathcal{G}_{train} = \{(S_i,M_k,T_{kj},F_{ijk} )| \\
          & \forall (S_i^\prime,M_k^\prime,T_{kj}^\prime,F_{ijk}^\prime ) \in \mathcal{G}_{test}, M_k \neq M_{k}^\prime \}
    \end{aligned}
    \end{equation}
    \item Method (c): Split all the brain signals $\mathcal{F}$ randomly with given ratio. 
    \begin{equation}
    \begin{aligned} \label{eq3}
         &\mathcal{G}_{train} = \{(S_i,M_k,T_{kj},F_{ijk} )| \\
          & \forall (S_i^\prime,M_k^\prime,T_{kj}^\prime,F_{ijk}^\prime ) \in \mathcal{G}_{test}, F_{ijk} \neq F_{ijk}^\prime \}
    \end{aligned}
    \end{equation}
    \item Method (d): Different from Method (c), it splits brain signals under each story randomly with given ratio, and union them to form the whole training and test set. 
    \item Method (e): Different from Method (d), it splits continuous brain signals under each story with given ratio, and union them to form the whole training and test set. 
\end{itemize}

To facilitate a thorough analysis of data leakage, we introduce the concept of \textit{brain signal leakage} and \textit{text stimuli leakage}.
Brain signal leakage refers to test subject's brain signal appears in training set. Text stimuli leakage refers to text segment in test set appears in the training set.
Formal definitions of two types of data leakage are given.
\begin{definition} 
    Brain signal leakage happens when 
    \begin{equation}\label{brain}
    \begin{aligned} 
        &\forall (S_i,M_k,T_{kj},F_{ijk}) \in \mathcal{G}_{train}, \\ 
        &\exists (S_i^\prime,M_k^\prime,T_{kj}^\prime,F_{ijk}^\prime) \in \mathcal{G}_{test}, S_i^\prime=S_i.
    \end{aligned}
    \end{equation}
\end{definition}
\begin{definition} 
    Text stimuli leakage happens when 
    \begin{equation} \label{text}
    \begin{aligned}
        &\forall (S_i,M_k,T_{kj},F_{ijk}) \in \mathcal{G}_{train}, \\ 
        &\exists (S_i^\prime,M_k^\prime,T_{kj}^\prime,F_{ijk}^\prime) \in \mathcal{G}_{test}, T_{kj}^\prime=T_{kj}.
    \end{aligned}
    \end{equation}
\end{definition}
Data leakage can be directly identified in graph $\mathcal{G}_\mathcal{D}$. As shown in Figure \ref{f2}, if edges connected to $S_i$ are of different colors, it indicates that brain signals of $S_i$ appears in both training set and test set, which leads to brain signal leakage. Similarly, if edges connected to $T_{kj}$ are of different colors, it suggests that text segment $T_{kj}$ appears in both training set and test set, which leads to text stimuli leakage.
\begin{table}
\centering
\resizebox{\columnwidth}{!}
{
\begin{tabular}{cccccc}
\toprule
fMRI / EEG   & Method(a) & Method(b) & Method(c) & Method(d) & Method(e)\\
\midrule
Brain Signal Leakage & \ding{55} / \ding{55} & \checkmark / \checkmark & \checkmark / \checkmark & \checkmark / \checkmark &\checkmark / NA \\
Text Stimuli Leakage & \checkmark / \checkmark & \ding{55} / \ding{55}&\checkmark / \checkmark &\checkmark / \checkmark &\checkmark / NA\\
\bottomrule
\end{tabular}
}
\caption{Data leakage in five different splitting methods applied to fMRI and EEG to text decoding separately.}
\label{fmrileak}
\vspace{-1mm}
\end{table}

As a result, in the scenario of EEG signals where $(S_i,M_k,T_{kj},F_{ijk})$ is viewed as a sample: Method (a) suffers from text stimuli leakage. Method (b) faces brain signal leakage. Method (c) is affected by leakage of both text stimuli and brain signals. Method (d) and (e) do not show any differences compared to method (c) in EEG-to-text decoding.
In fMRI-to-text decoding, continuous fMRI frames and text stimuli are concatenated to form one sample. $(S_i,M_k,T_{kj},F_{ijk})$ indicates the start point of one sample instead of the whole sample (recall Section \ref{2.3}). In this case, method (d) and (e) are different.
Similar to method (c), method (d) and (e) face both brain signal leakage and text stimuli leakage. But for method (e) the text stimuli is slight. It only happens in the overlapping part between training samples and test samples.
The situations of data leakage in different splitting methods are detailed in Table \ref{fmrileak}.

\begin{figure}
\centering
\includegraphics[width=0.49\textwidth]{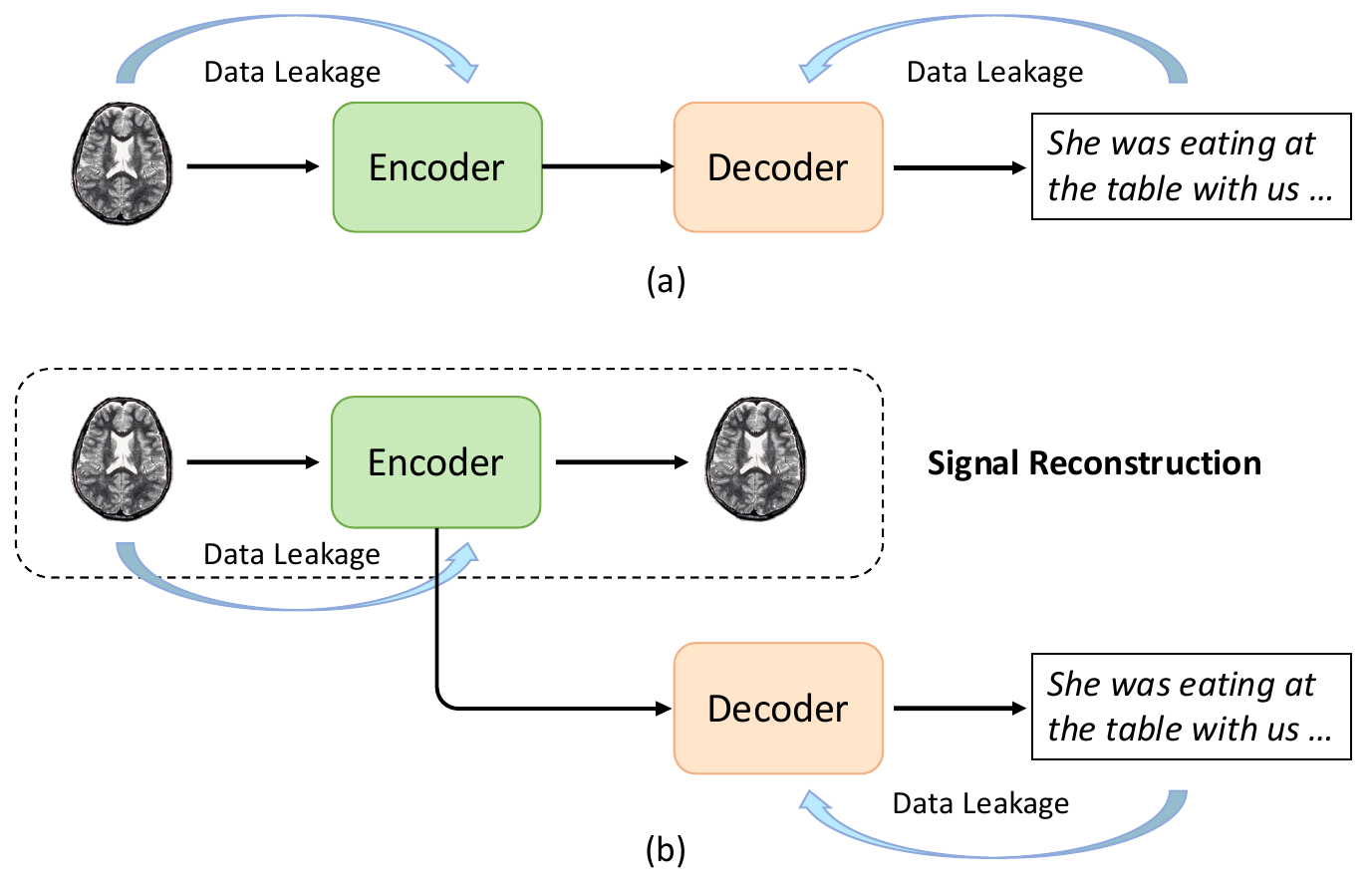}
\caption{General frameworks of current brain-to-text decoding models and how data leakage affect them.}
\label{f0}
\vspace{-1mm}
\end{figure}

\begin{figure*}
\centering
\includegraphics[width=\linewidth]{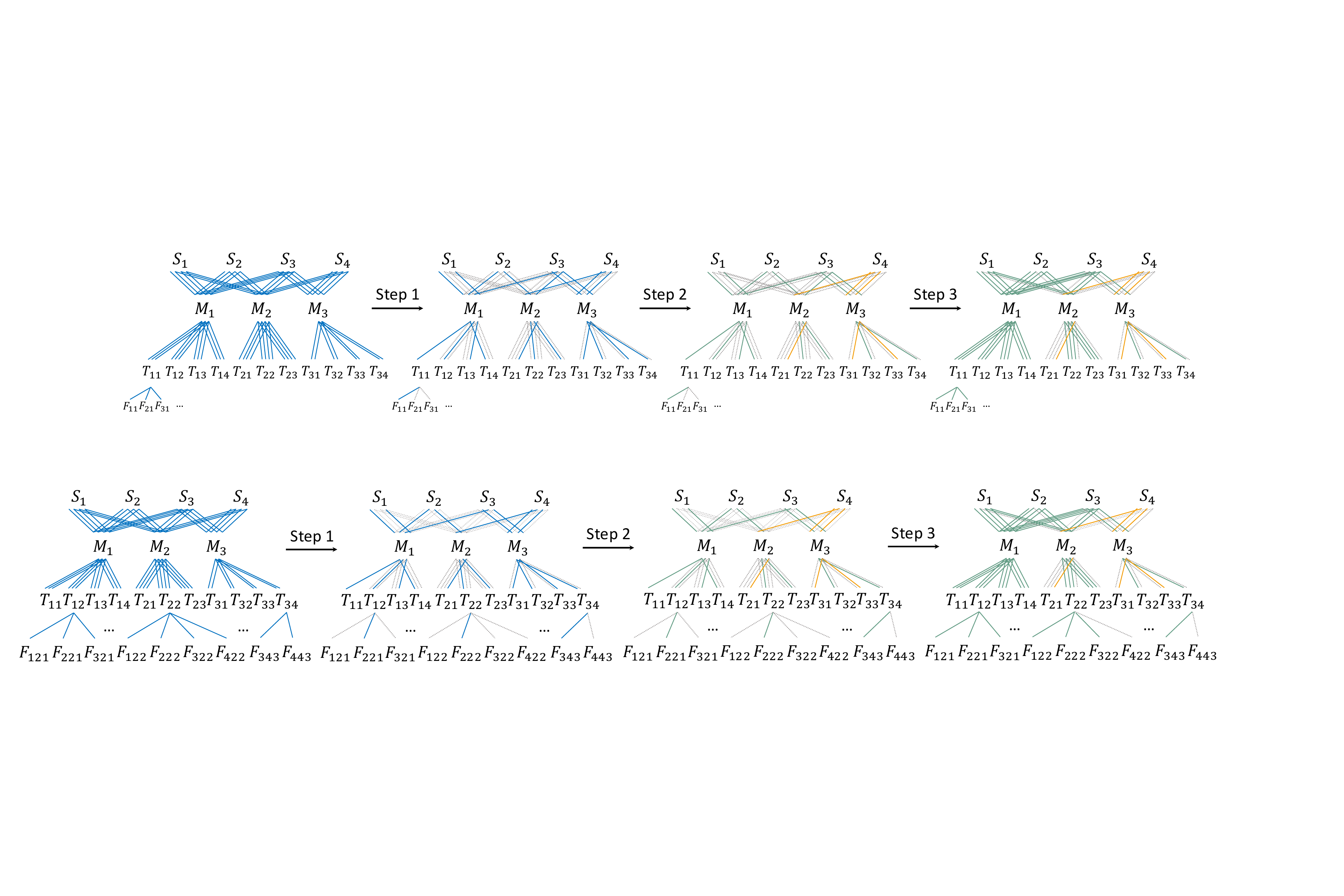}
\caption{The detailed steps of our proposed cross-subject data splitting method. 
 (Color printing is preferred.)}
\label{f4}
\end{figure*}


\subsection{Frameworks of Current Decoding Models and How Data Leakage Affect Them}
Current brain-to-text decoding models adopt an encoder-decoder framework, where the encoder is responsible for converting brain signals into low-dimensional representations and the decoder (usually Transformer-based) learns to map these representations to natural language. Two representative models EEG2Text \citep{DBLP:conf/aaai/WangJ22} and UniCoRN \citep{DBLP:conf/acl/XiZWL0023} are selected for investigating the affection of data leakage. As shown in Figure \ref{f0}(a), EEG2Text applies an end-to-end training manner. EEG feature sequence is first extracted by a multi-layer Transformer encoder and then converted to natural language with a pretrained BART \citep{DBLP:conf/acl/LewisLGGMLSZ20}. UniCoRN provides a unified framework for EEG and fMRI to text decoding. It follows a two-stage training manner as shown in Figure \ref{f0}(b). Take fMRI-to-text decoding for example, the encoder is first pre-trained with a brain signal reconstruction task to capture spatial and temporal feature via a 3D-CNN and multi-layer Transformer encoder module. Then BART \citep{DBLP:conf/acl/LewisLGGMLSZ20} is fine-tuned to translate fMRI representation into natural language.

Figure \ref{f0} illustrates how brain signal leakage and text stimuli leakage affect current encoder-decoder based brain-to-text decoding framework. For the encoder component, if subjects' brain signals in test set are mixed into training set, the encoder will become overfitted and fail to well represent unseen subjects' brain signals. For the decoder component, since it generates token by token in an auto-regressive manner, data leakage will cause the decoder to memorize seen text during the teacher-forcing training stage. The decoder will predict next token regardless of encoded brain signals.

\subsection{A Right Cross-Subject Splitting Method}
A right cross-subject splitting method is proposed to eliminate both brain signal leakage and text stimuli leakage.
\begin{equation} \label{mrule}
    \begin{aligned}
        &\mathcal{G}_{train} = \{(S_i,M_k,T_{kj},F_{ijk} )| \forall (S_i^\prime,M_k^\prime,\\
          &T_{kj}^\prime,F_{ijk}^\prime )   \in \mathcal{G}_{test}, S_i \neq S_{i}^\prime, T_{kj} \neq T_{kj}^\prime \}.
    \end{aligned}
\end{equation}
Given the differences of EEG and fMRI dataset, we address them separately and propose two data splitting methods.
In EEG dataset, $(S_i,M_k,T_{kj},F_{ijk} )$ forms one sample. As shown in Figure \ref{f4}, our proposed splitting method consists of three steps:
\begin{itemize}[topsep=0pt]
\setlength\itemsep{-0.5em}
    \item Step 1: Select $\sum_{k=1}^{K}l_k$ samples from $\mathcal{G}_\mathcal{D}$ and form a new graph $\mathcal{G}_\mathcal{D}^\prime$ that satisfies 
    \begin{equation}
    \begin{aligned}
        & \forall (S_i^\prime,M_k^\prime,T_{kj}^\prime,F_{ijk}^\prime), (S_i^{\prime\prime},M_k^{\prime\prime},T_{kj}^{\prime\prime},F_{ijk}^{\prime\prime})\\
        & \in \mathcal{G}_\mathcal{D}^\prime, T_{kj}^\prime \neq T_{kj}^{\prime\prime}.
    \end{aligned}
    \end{equation}
    \item Step 2: Split $\mathcal{G}_\mathcal{D}^\prime$ to $\mathcal{G}_{train}^\prime$ and $\mathcal{G}_{test}^\prime$ with a given ratio. The splitting should follow
    \begin{equation}
        \begin{aligned}
            &\mathcal{G}_{train}^\prime = \{(S_i,M_k,T_{kj},F_{ijk} )| \\
          & \forall (S_i^\prime,M_k^\prime,T_{kj}^\prime,F_{ijk}^\prime ) \in \mathcal{G}_{test}^\prime, S_i \neq S_{i}^\prime \},
        \end{aligned}
    \end{equation}
    \begin{equation}
        \begin{aligned}
            &\mathcal{G}_{test}^\prime = \{(S_i,M_k,T_{kj},F_{ijk})| \\
          & \forall (S_i^\prime,M_k^\prime,T_{kj}^\prime,F_{ijk}^\prime ) \in \mathcal{G}_{train}^\prime, S_i \neq S_{i}^\prime \}.
        \end{aligned}
    \end{equation}
    \item Step 3: Expand $\mathcal{G}_{train}^\prime$ and $\mathcal{G}_{test}^\prime$ with $\mathcal{G}_{train\_exp}^\prime$ and $\mathcal{G}_{test\_exp}^\prime$ separately. 
    \begin{equation}
    \begin{aligned}
        &\mathcal{G}_{train}^\prime\leftarrow \mathcal{G}_{train}^\prime \cup \mathcal{G}_{train\_exp}^\prime \\
        & \mathcal{G}_{test}^\prime\leftarrow \mathcal{G}_{test}^\prime \cup \mathcal{G}_{test\_exp}^\prime
    \end{aligned}
    \end{equation}
    where $\mathcal{G}_{train\_exp}^\prime$ and $\mathcal{G}_{test\_exp}^\prime$ are 
    \begin{equation}\label{13}
        \begin{aligned}
             \mathcal{G}_{train\_exp}^\prime &= \{(S_i,M_k,T_{kj},F_{ijk}) \in \mathcal{G}_\mathcal{D} | \\
             &S_i \in \mathcal{S}_{train}^\prime, 
             T_{kj} \in \mathcal{T}_{train}^\prime\},
        \end{aligned}
    \end{equation}
    \begin{equation}\label{14}
        \begin{aligned}
             \mathcal{G}_{test\_exp}^\prime &= \{(S_i,M_k,T_{kj},F_{ijk}) \in \mathcal{G}_\mathcal{D} | \\
             &S_i \in \mathcal{S}_{test}^\prime, 
             T_{kj} \in \mathcal{T}_{test}^\prime\}.
        \end{aligned}
    \end{equation}
\end{itemize}
$\mathcal{S}_{train}^\prime$, $\mathcal{T}_{train}^\prime$, $\mathcal{S}_{test}^\prime$, $\mathcal{T}_{test}^\prime$ indicate subject set, text segment set in $\mathcal{G}_{train}^\prime$ and subject set, text segment set in $\mathcal{G}_{test}^\prime$ respectively.

Some samples are discarded in our proposed splitting method, i.e. $\mathcal{G}_\mathcal{D} \neq \mathcal{G}_{train}^\prime \cup \mathcal{G}_{test}^\prime$. In Appendix \ref{proof}, we demonstrate that it is unavoidable for some samples to be discarded in order to satisfy the cross-subject data splitting criterion.

To fMRI dataset, continuous text segments and brain signals are concatenated to form one sample $(S_i,M_k,T_{kj}^*,F_{ijk}^*)$. If we follow the same splitting method as to EEG dataset, text stimuli leakage will happen in the overlapping part of two samples, when one sample is assigned to training set and the other is assigned to validation or test set. We propose a simple solution that achieves the balance between discarding as little data as possible while ensuring zero data leakage: Step 1 and Step 3 remain the same as splitting method for EEG dataset. In Step 2, $\mathcal{G}_{train}^\prime$ and $\mathcal{G}_{test}^\prime$ should follow
\begin{equation}
        \begin{aligned}
          &  \mathcal{G}_{train}^\prime = \{(S_i,M_k,T_{kj},F_{ijk} )| \forall (S_i^\prime,M_k^\prime, \\
          & T_{kj}^\prime,F_{ijk}^\prime ) \in \mathcal{G}_{test}^\prime, 
          S_i \neq S_{i}^\prime,M_k \neq M_{k}^\prime  \},
        \end{aligned}
    \end{equation}
    \begin{equation}
        \begin{aligned}
          &  \mathcal{G}_{test}^\prime = \{(S_i,M_k,T_{kj},F_{ijk} )| \forall (S_i^\prime,M_k^\prime, \\
          & T_{kj}^\prime,F_{ijk}^\prime ) \in \mathcal{G}_{train}^\prime, 
          S_i \neq S_{i}^\prime,M_k \neq M_{k}^\prime  \}.
        \end{aligned}
    \end{equation}

\section{Experimental Settings}
\subsection{Implementation Detail}
We test two SOTA cross-subject brain-to-text decoding models UniCoRN \citep{DBLP:conf/acl/XiZWL0023} and EEG2Text \citep{DBLP:conf/aaai/WangJ22} on fMRI dataset Narratives \citep{nastase2021narratives} and EEG dataset ZuCo \citep{hollenstein2018zuco}. Because the number of stories in ZuCo dataset is too small, and method (e) makes no difference to EEG as method (d), we only consider splitting method (a), (c), (d) for EEG.
We follow the same settings of UniCoRN and EEG2Text, except all the datasets are split to the ratio of 8:1:1 for fair comparison. Details are shown in Appendix \ref{sec:appendix}.

\subsection{Evaluation Metrics}
\paragraph{Data Leakage Metrics}
We design two novel evaluation metrics \textbf{Brain Signal Leakage Rate (BSLR)} and \textbf{Text Stimuli Leakage Rate (TSLR)} to quantify two types of data leakage. Note that the situation for validation set is the same as test set, so we only consider test set in experiments. BSLR indicates the average percentage of each subject's brain signals in test set appearing in training set, which could be formulated as
\begin{equation}
    \label{BILR}
    \frac{1}{N_{test}} \sum_{i=1}^{N_{test}} \min(1, \frac{|\{F_{ijk}|F_{ijk}\in (\mathcal{G}_{test}\cap \mathcal{G}_{train})\}|}{|\{F_{ijk}|F_{ijk} \in \mathcal{G}_{train}\}|})
\end{equation}
where $N_{test}$ stands for the total number of subjects in test set. $|\cdot|$ stands for the cardinality of a set. Function $\min(\cdot, \cdot)$ is applied to make sure for each subject the data leakage rate is less than one. 

The definition of TSLR is different for EEG signal and fMRI signal. Since $(S_i,M_k,T_{kj},F_{ijk} )$ indicates one sample in EEG dataset, definition of TSLR for EEG dataset is similar to BSLR, which measures the average percentage of certain text in test set appearing in training set.
\begin{equation}
    \label{EEG TSLR}
    \frac{1}{M_{test}} \sum_{j=1}^{M_{test}} \min(1, \frac{|\{T_{kj}|T_{kj}\in (\mathcal{G}_{test}\cap \mathcal{G}_{train})\}|}{|\{T_{kj}|T_{kj} \in \mathcal{G}_{train}\}|})
\end{equation}
where $M_{test}$ stands for the total number of text segments in test set.
\begin{table}
\centering
\resizebox{\columnwidth}{!}{
\begin{tabular}{ccccccc}
\toprule
    \textbf{Type} & \textbf{Method} & \textbf{Narratives} & \textbf{ZuCo} \\
\midrule
\multirow{6}{*}{BSLR$(\%)$}& (a)   & \textbf{0.00}$_{\pm0.00}$ & \textbf{0.00}$_{\pm0.00}$   \\
                      & (b)       & 9.67$_{\pm4.80}$ & /   \\
                      & (c)        & 12.50$_{\pm0.04}$ & 12.50$_{\pm0.03}$ \\
                      & (d)      & 12.80$_{\pm0.01}$ & 12.59$_{\pm0.02}$\\
                      & (e)   & 12.27$_{\pm0.01}$ & / \\
                      & (f)   & \textbf{0.00$_{\pm0.00}$} & \textbf{0.00$_{\pm0.00}$}   \\
\midrule
\multirow{6}{*}{TSLR$(\%)$} & (a)   & 100.00$_{\pm0.00}$ & 22.50$_{\pm1.31}$    \\
                      & (b)       & \textbf{0.00}$_{\pm0.00}$ & /   \\
                      & (c)        & 100.00$_{\pm0.00}$ & 13.07$_{\pm0.11}$     \\
                      & (d)    & 99.82$_{\pm0.17}$ & 12.88$_{\pm0.04}$ \\
                      & (e)   & 9.29$_{\pm0.06}$ & / \\
                      & (f)   & \textbf{0.00}$_{\pm0.00}$ & \textbf{0.00}$_{\pm0.00}$      \\
\bottomrule
\end{tabular}
}
\caption{Results of Brain Signal Leakage Rate (BSLR) and Text Stimuli Leakage Rate (TSLR). Lower is better.}
\label{tab:tab1}
\vspace{-3mm}
\end{table}
\begin{table*}
  \centering
    \begin{tabular}{ccccccc}
    \toprule
    \multirow{2}{*}{\textbf{Dataset}} & \multirow{2}{*}{\textbf{Model}} & \multirow{2}{*}{\textbf{Method}} & \multicolumn{4}{c}{\textbf{Original Test Set / Additional Test Set}}  \\
       \cmidrule(r){4-7}  &    &&   \multicolumn{1}{c}{BLEU-1} & \multicolumn{1}{c}{BLEU-2} & \multicolumn{1}{c}{BLEU-3} & \multicolumn{1}{c}{ROUGE1-F} \\
    \midrule
    \multirow{6}{*}{Narratives} & \multirow{6}{*}{UniCoRN} & \cellcolor{red!15}(a) & \cellcolor{red!15}49.56 / 18.43 & \cellcolor{red!15}30.49 / 1.25 & \cellcolor{red!15}21.07 / 0.00 & \cellcolor{red!15}40.65 / 16.38 \\
        && \cellcolor{green!15}(b) &\cellcolor{green!15}26.37 / 23.31  &\cellcolor{green!15}7.50 / 5.79  &\cellcolor{green!15}2.48 / 1.44     &\cellcolor{green!15}19.62 / 18.74 \\
        && \cellcolor{red!15}(c) &\cellcolor{red!15}50.24 / 16.96 & \cellcolor{red!15}30.83 / 0.09 & \cellcolor{red!15}21.23 / 0.00  &\cellcolor{red!15}41.01 / 15.12 \\
        && \cellcolor{red!15}(d) &\cellcolor{red!15}49.63 / 17.20 & \cellcolor{red!15}30.29 / 1.15 & \cellcolor{red!15}20.85 / 0.00  & \cellcolor{red!15}41.03 / 15.83 \\
        && \cellcolor{red!15}(e) & \cellcolor{red!15}28.94 / 21.79 &\cellcolor{red!15}9.39 / 4.62 &\cellcolor{red!15}4.07 / 1.19   & \cellcolor{red!15}19.49 / 18.78 \\
        && \cellcolor{green!15}(f) & \cellcolor{green!15}22.83 / 21.64  & \cellcolor{green!15}5.69 / 4.97 & \cellcolor{green!15}1.43 / 1.28  &  \cellcolor{green!15}19.04 / 18.45\\
    \midrule
    \multirow{8}{*}{ZuCo} & \multirow{4}{*}{UniCoRN}
        & \cellcolor{red!15}(a) & \cellcolor{red!15}58.09 / 18.54 & \cellcolor{red!15}49.23 / 1.31  & \cellcolor{red!15}43.23 / 0.00  &  \cellcolor{red!15}67.50 / 15.39\\
        &&\cellcolor{red!15}(c) & \cellcolor{red!15}52.30 / 18.38 &\cellcolor{red!15}42.89 / 1.03   &\cellcolor{red!15}36.80 / 0.00 & \cellcolor{red!15}67.29 /  15.25\\
        &&\cellcolor{red!15}(d) & \cellcolor{red!15}50.02 /  19.84 &\cellcolor{red!15}43.53 / 1.20   & \cellcolor{red!15}32.71 / 0.03  & \cellcolor{red!15}67.33 / 15.12  \\
        && \cellcolor{green!15}(f)& \cellcolor{green!15}23.32 / 22.89 & \cellcolor{green!15}7.78 / 7.46   & \cellcolor{green!15}3.01 / 2.75 & \cellcolor{green!15}17.92 / 17.63 \\
        \cmidrule{2-7}
        & \multirow{4}{*}{EEG2Text} & \cellcolor{red!15}(a) & \cellcolor{red!15}51.22 / 17.41 & \cellcolor{red!15}33.83 / 1.04 & \cellcolor{red!15}22.99 / 0.00  & \cellcolor{red!15}46.58 / 15.92 \\
        && \cellcolor{red!15}(c) & \cellcolor{red!15}53.83 / 17.38 & \cellcolor{red!15}38.99 / 0.84 & \cellcolor{red!15}29.57 / 0.00  & \cellcolor{red!15}53.56 / 16.07 \\
        && \cellcolor{red!15}(d) & \cellcolor{red!15}53.92/ 16.86 & \cellcolor{red!15}41.06 / 1.32  & \cellcolor{red!15}23.12 / 0.00   & \cellcolor{red!15}49.38 / 15.83\\
        && \cellcolor{green!15}(f)& \cellcolor{green!15}24.49 / 23.71 & \cellcolor{green!15}7.49 / 7.42  & \cellcolor{green!15}2.28 / 2.33 & \cellcolor{green!15}25.74 / 23.30\\
    \bottomrule
    \end{tabular}
      \caption{Performance of brain-to-text decoding models under different splitting methods on original test set and an additional test set. The \colorbox{green!15}{green mark} and \colorbox{red!15}{red mark} denotes a method without and with text stimuli leakage.}
  \label{addition}%
  \vspace{-3mm}
\end{table*}%
To fMRI dataset, continuous fMRI frames with corresponding text segments are concatenated as one sample. As a result, TSLR for fMRI signal is considered as the average percentage of the same text segments in test set appearing in training set, which is
\begin{equation}
    \label{fMRI TSLR1}
    \frac{1}{M_{test}} \sum_{j=1}^{M_{test}} \tau \frac{|\{T_{kj}|T_{kj}\in (\mathcal{G}_{test}\cap \mathcal{G}_{train})\}|}{|\mathcal{G}_{test}| \times L}
\end{equation}
where $\tau = 0$ if $\{T_{kj}|T_{kj}\in \mathcal{G}_{test}\cap \mathcal{G}_{train}\}=\emptyset$ else
\begin{equation}
    \label{fMRI TSLR2}
    \tau = \min (1, \frac{|\{T_{kj}|T_{kj}\in \mathcal{G}_{train}\}|}{|\{T_{kj}|T_{kj}\in (\mathcal{G}_{test}\cap \mathcal{G}_{train}\})|}).
\end{equation}

\paragraph{Decoding Performance Metrics}
BLEU \citep{DBLP:conf/acl/PapineniRWZ02} and ROUGE \citep{lin2004rouge} are applied to measure the decoding performance.
BLEU measures the n-gram overlap between decoded content and ground truth. 
ROUGE-N comparing the consistency of N-grams between the decoded content and the ground truth. 

\section{Experiments and Analysis}
We first quantify the data leakage condition of different methods with BSLR and TSLR metrics. Then we demonstrate the damage of data leakage on encoder side and decoder side. 
For model encoder, we analyze its validation loss under different splitting methods.
For model decoder, three experiment settings are applied: (1) An additional test set that ensures zero data leakage is left out as comparison to original test set. (2) The input brain signals are randomly shuffled. (3) Training original models with more epochs and smaller learning rate.

\subsection{Verification for Data Leakage}
Experiments on BSLR and TSLR are conducted four times with different seeds. The results in Table \ref{tab:tab1} are consistent with theoretical analysis. A value of zero in BSLR and TSLR demonstrate no brain signal leakage and text stimuli leakage, while higher values suggest more significant data leakage issues. Notably, only our method (f) prevents both brain signal leakage and text stimuli leakage.

\begin{figure}
\centering
\includegraphics[width=0.483\textwidth]{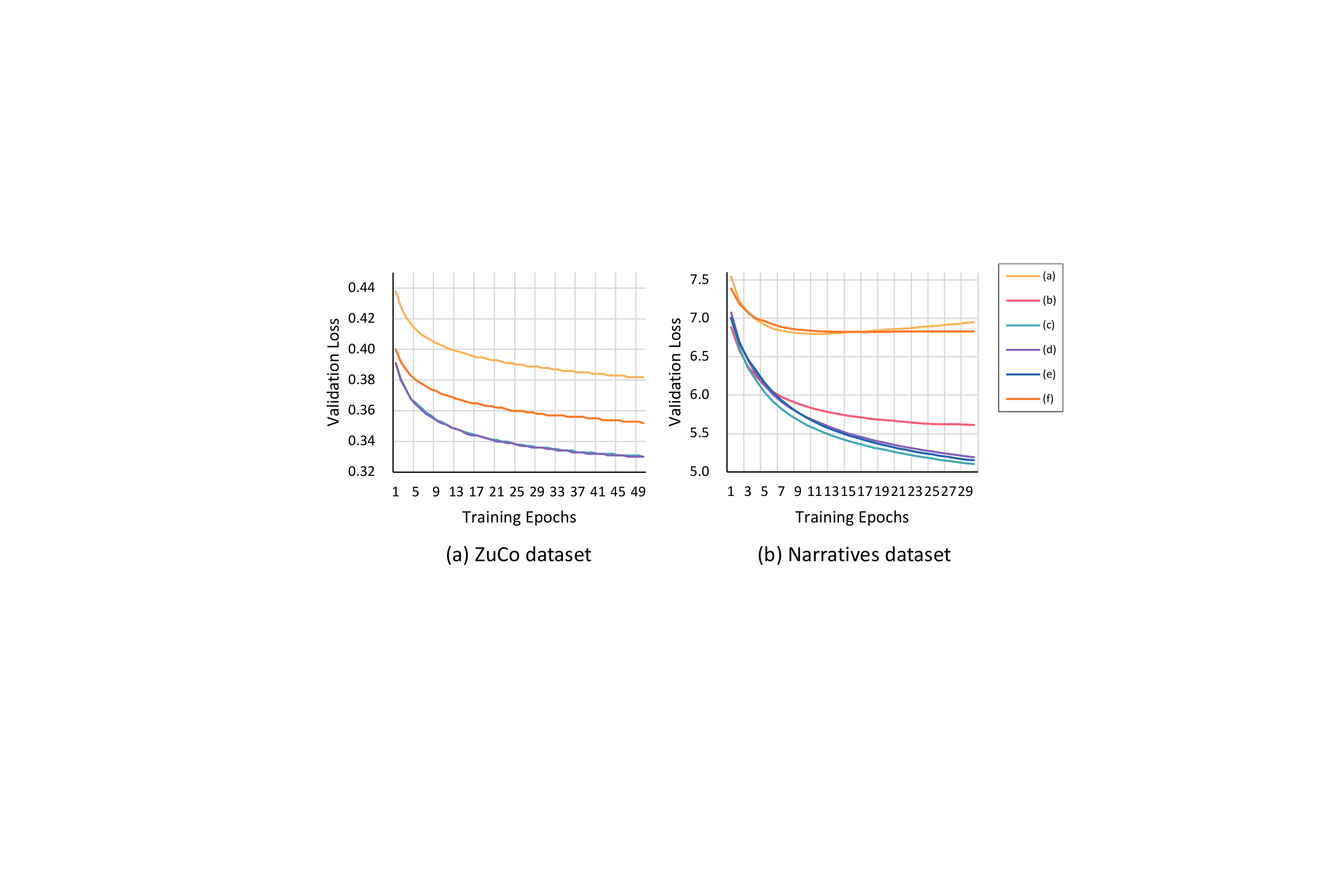}
\caption{Validation loss of encoder under different dataset splitting methods in two datasets.}
\label{f3}
\vspace{-3mm}
\end{figure}

\subsection{Damage of Data Leakage to Encoder}

\begin{table*}[t]
  \centering
    \begin{tabular}{ccccccc}
    \toprule
    \multirow{2}{*}{\textbf{Dataset}} & \multirow{2}{*}{\textbf{Model}} & \multirow{2}{*}{\textbf{Method}} & \multicolumn{4}{c}{\textbf{Ordered Input / Shuffled Input}}  \\
       \cmidrule(r){4-7}  &    &&   \multicolumn{1}{c}{BLEU-1} & \multicolumn{1}{c}{BLEU-2} & \multicolumn{1}{c}{BLEU-3} & \multicolumn{1}{c}{ROUGE1-F} \\
    \midrule
    \multirow{6}{*}{Narratives} & \multirow{6}{*}{UniCoRN} & \cellcolor{red!15}(a) & \cellcolor{red!15}49.56 / 47.39 & \cellcolor{red!15}30.49 / 28.95 & \cellcolor{red!15}21.07 / 18.40 & \cellcolor{red!15}40.65 / 35.12 \\
        && \cellcolor{green!15}(b) & \cellcolor{green!15}26.37 / 20.18  & \cellcolor{green!15}7.50 / 3.52  & \cellcolor{green!15}2.48 / 0.51     & \cellcolor{green!15}19.62 / 15.58 \\
        && \cellcolor{red!15}(c) & \cellcolor{red!15}50.24 / 48.48 & \cellcolor{red!15}30.83 / 30.21 & \cellcolor{red!15}21.23 / 19.39  &  \cellcolor{red!15}41.01 / 38.43 \\
        && \cellcolor{red!15}(d) & \cellcolor{red!15}49.63 / 50.21 & \cellcolor{red!15}30.29 / 32.18 & \cellcolor{red!15}20.85 / 21.46  & \cellcolor{red!15}41.03 / 41.69 \\
        && \cellcolor{red!15}(e) & \cellcolor{red!15}28.94 / 24.84 & \cellcolor{red!15}9.39 / 6.56 & \cellcolor{red!15}4.07 / 2.04   & \cellcolor{red!15}19.49 / 17.90 \\
        && \cellcolor{green!15}(f) & \cellcolor{green!15}22.83 / 18.21  & \cellcolor{green!15}5.69 / 2.47 & \cellcolor{green!15}1.43 / 0.22  &  \cellcolor{green!15}19.04 / 16.83\\
    \midrule
    \multirow{8}{*}{ZuCo} & \multirow{4}{*}{UniCoRN}
        & \cellcolor{red!15}(a) & \cellcolor{red!15}58.09 / 59.23   & \cellcolor{red!15}49.23 / 51.35 & \cellcolor{red!15}43.23 / 44.27  &  \cellcolor{red!15}67.50 / 68.93  \\
        && \cellcolor{red!15}(c) & \cellcolor{red!15}52.30 / 50.24  & \cellcolor{red!15}42.89 / 37.96 & \cellcolor{red!15}36.80 / 30.21 & \cellcolor{red!15}67.29 /  63.43\\
        && \cellcolor{red!15}(d) & \cellcolor{red!15}50.02 / 51.12 & \cellcolor{red!15}43.53 / 40.85   & \cellcolor{red!15}32.71 / 28.24 & \cellcolor{red!15}67.33 / 64.88 \\
        && \cellcolor{green!15}(f) & \cellcolor{green!15}23.32 / 19.38 & \cellcolor{green!15}7.78 / 2.51  & \cellcolor{green!15}3.01 / 0.00 & \cellcolor{green!15}17.92 / 15.21 \\
        \cmidrule{2-7}
        & \multirow{4}{*}{EEG2Text} & \cellcolor{red!15}(a) & \cellcolor{red!15}51.22 / 50.63 & \cellcolor{red!15}33.83 / 32.19 & \cellcolor{red!15}22.99 / 20.63 & \cellcolor{red!15}46.58 / 44.70 \\
        && \cellcolor{red!15}(c) & \cellcolor{red!15}53.83 / 50.33 & \cellcolor{red!15}38.99 / 33.42 & \cellcolor{red!15}29.57 / 23.19  & \cellcolor{red!15}53.56 / 48.78 \\
        && \cellcolor{red!15}(d) & \cellcolor{red!15}53.92 / 51.46 & \cellcolor{red!15}41.06 / 35.87  & \cellcolor{red!15}23.12 / 24.75  & \cellcolor{red!15}49.38 / 47.42\\
        && \cellcolor{green!15}(f) & \cellcolor{green!15}24.49 / 18.72 & \cellcolor{green!15}7.49 / 2.01 & \cellcolor{green!15}2.28 / 0.00 & \cellcolor{green!15}25.74 / 15.36\\
    \bottomrule
    \end{tabular}
      \caption{Performance of brain-to-text decoding models under different splitting methods with ordered brain signals and randomly shuffled brain signals as model input respectively. The \colorbox{green!15}{green mark} and \colorbox{red!15}{red mark} denotes a method without and with text stimuli leakage correspondingly.}
  \label{shuffle}%
  \vspace{-3mm}
\end{table*}%

\begin{table}
  \centering
  \resizebox{\columnwidth}{!}{
    \begin{tabular}{ccccccccc}
    \toprule
    \multirow{2}{*}{\textbf{Dataset}} & \multirow{2}{*}{\textbf{Model}} & \multicolumn{4}{c}{\textbf{BLEU-N ($\%$)}} & \multicolumn{3}{c}{\textbf{ROUGE-1 ($\%$)}} \\
       \cmidrule(r){3-6}  \cmidrule(r){7-9}&      & \multicolumn{1}{c}{$N=1$} & \multicolumn{1}{c}{$N=2$} & \multicolumn{1}{c}{$N=3$} & \multicolumn{1}{c}{$N=4$} & \multicolumn{1}{c}{$R$} & \multicolumn{1}{c}{$P$} & \multicolumn{1}{c}{$F$} \\
    \midrule
    Narratives & UniCoRN & \textbf{22.83}  & \textbf{5.69} & \textbf{1.43} & \textbf{0.48} & \textbf{15.55} & \textbf{24.80} & \textbf{19.04}\\
    \midrule
    \multirow{2}{*}{ZuCo} & UniCoRN & 23.32 & \textbf{7.78} & \textbf{3.01} & \textbf{1.09}  & 18.47 & 20.00 & 17.92 \\
        & EEG2Text & \textbf{24.49} & 7.49 & 2.28 & 0.62 & \textbf{23.98} & \textbf{23.95} & \textbf{25.74} \\
    \bottomrule
    \end{tabular}
    }
    \caption{A fair benchmark for evaluating the performance of cross-subject brain-to-text decoding models.}
      \label{tab:tab4}%
    \vspace{-3mm}
\end{table}%

Evaluating the encoder independently can be challenging in an end-to-end training scenario. Therefore, we primarily focus on a pre-trained encoder. Validation loss is applied to measure data leakage, as a proper evaluation index of encoder's representation ability is missing.
The validation loss of encoder under different data splitting methods is shown in Figure \ref{f3}. For fMRI data, the presence of brain signal leakage causes the validation loss of methods (b), (c), (d), and (e) to continuously decrease even over extended training epochs, which indicates the encoder is actually overfitting and its representation ability is degrading. In contrast, with methods (a) and (f) that are not affected by brain signal leakage, the validation loss quickly increases after reaching its minimum within a few epochs. For EEG, we find validation loss keeps dropping for all methods even with very long training epochs, regardless of brain signal leakage or not. We think the poor spatial resolution of EEG signal might lead to this phenomenon.

\subsection{Damage of Data Leakage to Decoder}
\paragraph{Evaluation on Additional Test Set}
An additional test set that ensures zero data leakage is left out to evaluate the actual performance of brain-to-text decoding models. If the original test set is correctly split, its decoding result should be similar to that of the additional test set.
From Table \ref{addition}, we observe that the decoding model tends to overfit when text stimuli leakage occurs, as seen in methods (a), (c), (d), and (e) in Narratives, and methods (a) and (c) in ZuCo. The BLEU and ROUGE score is significantly lower in the additional test set. 
While in our proposed splitting method (f), the decoding performance of original and additional test set are similar.
We also notice that methods with a high Text Stimuli Leakage Rate (TSLR), such as method (a) in Narratives, exhibit more overfitting compared to methods with a low TSLR, like method (e).
\paragraph{Shuffle Input Brain Signals}
We conduct a chance-level experiment to investigate whether decoding models learn language reconstruction from brain signals. Specifically, the input brain signals are randomly shuffled. Decoding performance in test set is expected to be very poor if text stimuli leakage does not happen, as the shuffled input is considered as noise. However, if text stimuli in test set leaks into training set, the model will simply memorize seen text and the decoding performance is not supposed to be affected.

Results are presented in Table \ref{shuffle}. For fMRI, we find the decoding performance of models under splitting method (a), (c), and (d) remain the same no matter the input is ordered or shuffled. Similar phenomenon is also observed in EEG dataset when it comes to splitting method (a), (c), (d).
But in splitting method without text stimuli leakage, model performance with shuffled input drops significantly.
This experiment demonstrates that the brain-to-text decoding task become meaningless when text stimuli leakage exists, as the Transformer block is capable of generating text that was previously encountered during the training phase.
\paragraph{Longer Training Epochs with Smaller Learning Rate}
According to fundamental machine learning principle, model performance in test set will first increase and then drop as the training proceeds. In this experiment, we try training models under different splitting methods with longer training epochs and smaller learning rate. If text stimuli leakage happens, the model is overfitting and its performance is supposed to keep increasing.

Results and detailed analysis are presented in Appendix \ref{longepoch}. In conclusion, the model's performance on test set continues to improve when text stimuli leakage happens, confirming that such leakage results in significant overfitting in decoding models.

\subsection{A Fair Benchmark}
We re-evaluate two SOTA models for brain-to-text decoding under our cross-subject data splitting method and release a fair benchmark. UniCoRN is tested for both fMRI and EEG decoding, EEG2Text model is tested for EEG decoding. The results are listed in Table \ref{tab:tab4}. For EEG dataset, UniCoRN achieves higher results in BLEU-2,3,4 while EEG2Text is better in BLEU-1 and ROUGE-1.

\section{Conclusion and Discussion}
In this paper, we evidence that all current dataset splitting methods for cross-subject brain-to-text decoding have data leakage problem through theoretical analysis and experiments. Such data leakage leads to model overfitting and largely exaggerates model performance, rendering model evaluation meaningless. 
To fix this issue, we propose a right cross-subject data splitting method. Current SOTA models are re-evaluated for further researches.

It's essential to realize the false promise of current SOTA methods.
This inspires future researches to design more general and accurate models for brain-to-text decoding, under the right cross-subject dataset splitting method.
It should be noted that we don't seek to propose any modifications to improve current models in the scope of this paper.
We mainly focus on revealing the false dataset splitting method and its detrimental impact on cross-subject brain-to-text decoding research.

\section*{Limitations}
The limitations of this work include three aspects: (1) Although our splitting method can be applied to any natural language comprehension cognitive dataset, we only analyze cross-subject data splitting methods in fMRI and EEG dataset. We leave the investigation of other cognitive signals (e.g. ECoG, MEG, etc.) to future work. (2) Our proposed dataset splitting method meets the above requirements at the expense of discarding some data in the dataset. We recommend future datasets in this domain follow these guidelines. The division of the training set, validation set, and test set should be provided when the dataset is released. Besides, we suggest hiring new subjects with unique stimuli for the validation set and test set, which is good for testing the generalization ability of models without loss of data. (3) During experiments we find existing models rely more on a strong auto-regressive decoder to achieve good generation quality. The encoder is of limited use in all SOTA models. And we also notice in experiments that the encoder of EEG2Text keeps overfitting whether with or without brain signal leakage. We leave it as future research.

\section*{Ethics Statement}
In this paper, we introduce a new dataset splitting method to avoid data leakage for decoding brain signals to text task. Experiments are conducted on the publicly accessible cognitive datasets ``Narratives'' and ZuCo1.0 with the authorization from their respective maintainers. Both datasets have been de-identified by dataset providers and used for researches only.

\section*{Acknowledgements}
This research is supported by the National Natural Science Foundation of China (No.62476127),  the Natural Science Foundation of Jiangsu Province (No.BK20242039), the Basic Research Program of the Bureau of Science and Technology (ILF24001), the Meituan Research Fund (No.APAP202507180115), the Scientific Research Starting Foundation of Nanjing University of Aeronautics and Astronautics (No.YQR21022), and the High Performance Computing Platform of Nanjing University of Aeronautics and Astronautics.

\bibliography{custom}
\bibliographystyle{acl_natbib}

\appendix

\section{Related Work}
\paragraph{Brain Signal}
Brain signals can be classified into three categories: invasive, partially invasive, and non-invasive according to how close electrodes get to brain tissue. In this paper, we mainly focus on non-invasive signals EEG and fMRI. EEG signal is electrogram of the spontaneous electrical activity of the brain, with frequencies ranging from 1 Hz to 30 Hz. EEG is of high temporal resolution and relatively tolerant of subject movement, but its spatial resolution is low and it can't display active areas of the brain directly. fMRI measures brain activity by detecting changes of blood flow. Blood flow of a specific region increases when this brain area is in use. The spatial resolution of fMRI is measured by the size of voxel, which is a three-dimensional rectangular cuboid ranging from 3mm to 5mm \citep{vouloumanos2001detection,noppeney2004fmri}. Unlike EEG which samples brain signals continuously, fMRI samples based on a fixed time interval named TR, usually at second level.

\paragraph{Brain-to-text Decoding}
Previous research on brain-to-text decoding \citep{herff2015brain,anumanchipalli2019speech,zou2021towards,moses2021neuroprosthesis,D_fossez_2023} mainly focused on word-level decoding in a restricted vocabulary with hundreds of words \citep{panachakel2021decoding}. These models typically apply recurrent neural network or long short-term memory \citep{DBLP:journals/neco/HochreiterS97} network to build mapping between brain signals and words in vocabulary. Despite relatively good accuracy, these methods fail to generalize to unseen words. Some progress \citep{DBLP:conf/aaai/SunWZZ19} has been made by expanding word-level decoding to sentence-level through encoder-decoder framework or using less noisy ECoG data \citep{burle2015spatial,anumanchipalli2019speech}. However, these models struggle to generate accurate and fluent sentences limited by decoder ability. \citet{DBLP:conf/aaai/WangJ22} introduced the first open vocabulary EEG-to-text decoding model by leveraging the power of pre-trained language models. \citet{DBLP:conf/acl/XiZWL0023} improved the model design and proposed a unified framework for decoding both fMRI and EEG signals.


\section{Implementation Details}
\label{sec:appendix}
We apply the ``Narratives'' \citep{nastase2021narratives} dataset for fMRI-to-text decoding and the ZuCo \citep{hollenstein2018zuco} dataset for EEG-to-text decoding in experiments. The ``Narratives'' dataset contains fMRI data from 345 subjects listening to 27 diverse stories. Since the data collection process involves different machines, we only consider fMRI data with $64 \times 64 \times 27$ voxels. The ZuCo dataset includes 12 healthy adult native English speakers reading English text for 4 to 6 hours. It contains simultaneous EEG and Eye-tracking data. The reading tasks include Normal Reading (NR) and Task-specific Reading (TSR) extracted from movie views and Wikipedia. Both datasets are split into training, validation, and test set with a ratio of 80$\%$, 10$\%$, 10$\%$ in all experiments.

We perform the same filtering steps to ``Narratives'' dataset as UniCoRN paper \citep{DBLP:conf/acl/XiZWL0023} and the same filtering steps to ZuCo1.0 as EEG2Text paper \citep{DBLP:conf/aaai/WangJ22}. In BSLR and TSLR calculation, the number of four different seeds are set as $1,2,3,4$ respectively. In signal reconstruction task for encoder of UniCoRN, the batch size of EEG and fMRI data is 512 and 320 respectively. The learning rate is set as 1e-4 and 1e-3 separately as the author claimed in the original paper. In the fair benchmark, for fMRI data, encoder of UniCoRN is trained through 1e-4 learning rate and decaying to 1e-6 finally for 30 training epochs. Decoder is trained through 1e-4 learning rate and decaying to 1e-6 finally for 10 training epochs with 90 batch size. Sample length $L$ is set as 10 for all experiments related to fMRI. For EEG data, EEG2Text model is trained with 1e-6 learning rate for 80 epochs. UniCoRN model is trained with the same settings as fMRI data.

\section{Cross-Subject Data Splitting in Practice}
\label{sec:appendix2}
We present the pseudo-code of two dataset splitting methods for EEG and fMRI signal. We only consider a bipartite graph $\mathcal{G}_1=(\mathcal{U},\mathcal{V},\mathcal{E})$ instead of a $4$\textit{-partite} graph in real practice.
For EEG signal, $\mathcal{U} = \{S_i\}_{i=1}^{N}$, $\mathcal{V} = \{T_j\}_{j=1}^{M}$. While for fMRI signal, $\mathcal{U} = \{S_i\}_{i=1}^{N}$, $\mathcal{V} = \{M_k\}_{k=1}^{K}$.
$\mathcal{E}$ is the edge between node in $\mathcal{U}$ and node in $\mathcal{V}$. $N,M,K$ indicate the total number of subjects, text segments and stories. We assert $M > N$ for EEG dataset and $K < N$ for fMRI dataset, so $e=(u,v) \in \mathcal{E}$ exists for every $v \in \mathcal{V}$, as each text segment or story is listened by at least one subject. As shown in step 1 of Figure \ref{f4}, first we pick one edge for each node $v \in \mathcal{V}$ and build a new bipartite graph $\mathcal{G}_2=(\mathcal{U},\mathcal{V},\mathcal{E}^\prime)$. Then following step 2, we split graph $\mathcal{G}_2$ by subject $\mathcal{U}$ with the given splitting ratio and form three disjoint graphs $\mathcal{G}_{train}, \mathcal{G}_{val}, \mathcal{G}_{test}$. In step 3, we extend each graph $\mathcal{G}_{train}, \mathcal{G}_{val}, \mathcal{G}_{test}$ by adding edges without data leakage.

The main difference of splitting methods for EEG and fMRI lies in how $\mathcal{G}_2$ is generated. We always choose the side with fewer nodes in bipartite graph $\mathcal{G}_1$ to generate $\mathcal{G}_2$. Specifically, in Algorithm \ref{al1} where we assert $|\mathcal{U}| < |\mathcal{V}|$, the adjacency matrix is initialized as $M \times N$. In Algorithm \ref{al2} where $|\mathcal{V}| < |\mathcal{U}|$, the adjacency matrix is initialized as $N \times K$. All assertions are based on real cognitive datasets.
One more thing to notice is that in Line 14 of both pseudo-code, the loop indicates extending training set, validation set, and test set respectively. So the names of variable should be alternated in the repeat loop and the displayed part in pseudo-code is a case example of extending training set. We write it in this way for simplicity of expression.

\section{Supplementary Proof} \label{proof}
\begin{definition} \label{multi}
    An directed \textit{multigraph} $\mathcal{G}$ is a type of graph which is permitted to have multiple edges between two vertices. When the edges own identity,  $\mathcal{G}$ can be written as $\mathcal{G}=(\mathcal{V},\mathcal{E},f)$, where $f: \mathcal{E} \rightarrow \mathcal{V}\times \mathcal{V}$ is an incidence function that maps each edge to a pair of vertices.
\end{definition} 

\begin{definition} \label{k}
    A $k$\textit{-partite graph} $\mathcal{G}$ is a type of graph that can be divided into $k$ distinct independent sets such that no two vertices in the same set are connected. $\mathcal{G}=(\mathcal{V},\mathcal{E})$, where $\mathcal{V}=\mathcal{V}_1 \cup \mathcal{V}_2 \cup \cdots \cup \mathcal{V}_k$ and $\forall i \neq j, \mathcal{V}_i \cap \mathcal{V}_j = \emptyset$.
\end{definition}

\begin{notation}
    $\otimes$ is a Cartesian product-like operator.
     $X \otimes Y = \{(x,y)|x\in X, y\in Y,$ there exists relationship between $x$ and $y$ in dataset$\}$. It's designed to describe the connectivity among $\mathcal{S}, \mathcal{M}, \mathcal{T}, \mathcal{F}$.
     For example, edges in $\mathcal{S}\otimes\mathcal{M}$ indicates certain subjects are stimulated by certain stories as described in dataset.
\end{notation}

\begin{definition} \label{rule}
    The training set for cross-subject brain-to-text decoding should be formatted in $\mathcal{G}_{train} = \mathcal{S}_{train} \otimes \mathcal{M} \otimes \mathcal{T}_{train}\otimes \mathcal{F}_{train}$, where $\mathcal{S}_{train} = \{ S_i | \forall S_i^\prime \in \mathcal{S}_{test}, S_i \neq S_i^\prime \}$;
    $\mathcal{F}_{train}=\{F_{ijk}|i\in I\}$, $I=\{i|\forall j, \forall k, F_{ijk}\notin \mathcal{F}_{test}\}$;
    $\mathcal{T}_{train} = \{ T_{kj} | \forall T_{kj}^\prime \in \mathcal{T}_{test}, T_{kj} \neq T_{kj}^\prime \}$.
\end{definition}

\paragraph{Why a method without brain signal leakage and text stimuli leakage must satisfy cross-subject brain-to-text decoding criterion} Training set $\mathcal{G}_{train}$ without brain signal leakage and text stimuli leakage is formatted in
\begin{equation}
    \begin{aligned}
        \mathcal{G}_{train} &= \{(S_i,M_k,T_{kj},F_{ijk})| \\
          & \forall (S_i^\prime,M_k^\prime,T_{kj}^\prime,F_{ijk}^\prime ) \in \mathcal{G}_{test}, \\
            & S_i \neq S_{i}^\prime, T_{kj}\neq T_{kj}^\prime\} \\
            &= \mathcal{S}_{train} \otimes \mathcal{M} \otimes \mathcal{T}_{train}\otimes \mathcal{F} 
    \end{aligned}
\end{equation}
where $\mathcal{S}_{train} = \{ S_i | \forall S_i^\prime \in \mathcal{S}_{test}, S_i \neq S_i^\prime \},\mathcal{T}_{train} = \{ T_{kj} | \forall T_{kj}^\prime \in \mathcal{T}_{test}, T_{kj} \neq T_{kj}^\prime \}$. Since $F_{ijk} \in \mathcal{F}$ indicates brain signal of subject $S_i$ stimulated by text segment $T_{kj}$, and given the definition of operator $\otimes$, $\mathcal{F}$ is determined when $\mathcal{S}$ and $\mathcal{T}$ are specified, which is 
\begin{equation}
    \begin{aligned}
        \mathcal{F}&=\{F_{ijk} | i \in I, kj \in J\}, \\
        I &= \{i| S_i \in \mathcal{S}_{train}\},\\
        J &= \{kj|T_{kj} \in \mathcal{T}_{train}\}.
    \end{aligned}
\end{equation}
$\mathcal{F}$ can also be written as $\mathcal{F}=\{F_{ijk}|i\in I\}$, $I=\{i|\forall j, \forall k, F_{ijk}\notin \mathcal{F}_{test}\}$, which is equal to Definition \ref{rule}.
\paragraph{Why the proposed splitting method satisfy zero data leakage}
Take the splitting method for EEG signal as example, the training set and test set after step 1 and step 2 already satisfy
\begin{equation}
        \begin{aligned}
            &\mathcal{G}_{train} = \{(S_i,M_k,T_{kj},F_{ijk} )| \forall (S_i^\prime,M_k^\prime,\\
          &T_{kj}^\prime,F_{ijk}^\prime )   \in \mathcal{G}_{test}, S_i \neq S_{i}^\prime, T_{kj} \neq T_{kj}^\prime \}\\
        \end{aligned}
    \end{equation}
    \begin{equation}
        \begin{aligned}
            &\mathcal{G}_{test} = \{(S_i,M_k,T_{kj},F_{ijk} )| \forall (S_i^\prime,M_k^\prime,\\
          &T_{kj}^\prime,F_{ijk}^\prime )   \in \mathcal{G}_{train}, S_i \neq S_{i}^\prime, T_{kj} \neq T_{kj}^\prime \}\\
        \end{aligned}
    \end{equation}
So we only need to prove expanded graph $\mathcal{G}_{train\_exp}^\prime$ and $\mathcal{G}_{test\_exp}^\prime$ satisfy zero data leakage, which is obvious from Equation \ref{13} and \ref{14}.
\paragraph{Why we must discard samples to ensure no data leakage}
If $\mathcal{G}_{train}\cup \mathcal{G}_{test}=\mathcal{G}_\mathcal{D}$, suppose $\forall (S_i,M_k,T_{kj},F_{ijk}) \in \mathcal{G}_{train}$, $(S_i^\prime,M_k^\prime,T_{kj}^\prime,F_{ijk}^\prime) \in \mathcal{G}_{test}, S_i \neq S_i^\prime, T_{kj} \neq T_{kj}^\prime$. For $f(\mathcal{E})=(M_k,T_{kj})$, $f(\mathcal{E}^\prime)=(M_k^\prime,T_{kj}^\prime)$, $T_{kj} \neq T_{kj}^\prime$:
\begin{itemize}
    \item If $M_k = M_k^\prime$, then there must exist a subject $S_i=S_i^\prime$ such that he is stimulated by the whole stories.
    \item If $M_k \neq M_k^\prime$, then there must exist a subject $S_i=S_i^\prime$ such that he is stimulated by two different stories.
\end{itemize}
As a result, if $\mathcal{G}_{train}\cup \mathcal{G}_{test}=\mathcal{G}_\mathcal{D}$, then $\exists (S_i,M_k,T_{kj},F_{ijk}) \in \mathcal{G}_{train}$, $(S_i^\prime,M_k^\prime,T_{kj}^\prime,F_{ijk}^\prime) \in \mathcal{G}_{test}$, $s.t.$ $S_i=S_i^\prime$ or $T_{kj}=T_{kj}^\prime$. Some samples must be discarded to ensure no data leakage.

\begin{table*}[t]
  \centering
    \begin{tabular}{ccccccc}
    \toprule
    \multirow{2}{*}{\textbf{Dataset}} & \multirow{2}{*}{\textbf{Model}} & \multirow{2}{*}{\textbf{Method}} & \multicolumn{4}{c}{\textbf{With Teacher-Forcing / Without Teacher-Forcing}}  \\
       \cmidrule(r){4-7}  &    &&   \multicolumn{1}{c}{BLEU-1} & \multicolumn{1}{c}{BLEU-2} & \multicolumn{1}{c}{BLEU-3} & \multicolumn{1}{c}{ROUGE1-F} \\
    \midrule
    \multirow{6}{*}{Narratives} & \multirow{6}{*}{UniCoRN} & \cellcolor{red!15}(a) & \cellcolor{red!15}49.56 / 33.82 & \cellcolor{red!15}30.49 / 9.44 & \cellcolor{red!15}21.07 / 4.89 & \cellcolor{red!15}40.65 / 21.87 \\
        && \cellcolor{green!15}(b) & \cellcolor{green!15}26.37 / 19.68  & \cellcolor{green!15}7.50 / 5.50  & \cellcolor{green!15}2.48 / 1.76     & \cellcolor{green!15}19.62 / 16.86 \\
        && \cellcolor{red!15}(c) & \cellcolor{red!15}50.24 / 29.87 & \cellcolor{red!15}30.83 / 8.35 & \cellcolor{red!15}21.23 / 3.21  &  \cellcolor{red!15}41.01 / 19.03 \\
        && \cellcolor{red!15}(d) & \cellcolor{red!15}49.63 / 30.20 & \cellcolor{red!15}30.29 / 8.78 & \cellcolor{red!15}20.85 / 3.65  & \cellcolor{red!15}41.03 / 19.42 \\
        && \cellcolor{red!15}(e) & \cellcolor{red!15}28.94 / 21.46 & \cellcolor{red!15}9.39 / 6.24 & \cellcolor{red!15}4.07 / 1.83   & \cellcolor{red!15}19.49 / 17.39 \\
        && \cellcolor{green!15}(f) & \cellcolor{green!15}22.83 / 16.85  & \cellcolor{green!15}5.69 / 4.24 & \cellcolor{green!15}1.43 / 0.65  &  \cellcolor{green!15}19.04 / 15.34\\
    \midrule
    \multirow{8}{*}{ZuCo} & \multirow{4}{*}{UniCoRN}
        & \cellcolor{red!15}(a) & \cellcolor{red!15}58.09 / 19.47   & \cellcolor{red!15}49.23 / 7.69 & \cellcolor{red!15}43.23 / 2.97  &  \cellcolor{red!15}67.50 / 17.25  \\
        && \cellcolor{red!15}(c) & \cellcolor{red!15}52.30 / 19.70  & \cellcolor{red!15}42.89 / 7.54 & \cellcolor{red!15}36.80 / 2.93 & \cellcolor{red!15}67.29 /  17.37\\
        && \cellcolor{red!15}(d) & \cellcolor{red!15}50.02 / 22.02 & \cellcolor{red!15}43.53 / 8.28   & \cellcolor{red!15}32.71 / 3.15 & \cellcolor{red!15}67.33 / 18.33 \\
        && \cellcolor{green!15}(f) & \cellcolor{green!15}23.32 / 14.02 & \cellcolor{green!15}7.78 / 2.57  & \cellcolor{green!15}3.01 / 0.82 & \cellcolor{green!15}17.92 / 11.95 \\
        \cmidrule{2-7}
        & \multirow{4}{*}{EEG2Text} & \cellcolor{red!15}(a) & \cellcolor{red!15}51.22 / 21.99 & \cellcolor{red!15}33.83 / 7.42 & \cellcolor{red!15}22.99 / 2.94 & \cellcolor{red!15}46.58 / 17.78 \\
        && \cellcolor{red!15}(c) & \cellcolor{red!15}53.83 / 20.41 & \cellcolor{red!15}38.99 / 7.25 & \cellcolor{red!15}29.57 / 2.48  & \cellcolor{red!15}53.56 / 17.32 \\
        && \cellcolor{red!15}(d) & \cellcolor{red!15}53.92 / 19.80 & \cellcolor{red!15}41.06 / 7.46  & \cellcolor{red!15}23.12 / 3.06  & \cellcolor{red!15}49.38 / 17.29\\
        && \cellcolor{green!15}(f) & \cellcolor{green!15}24.49 / 13.52 & \cellcolor{green!15}7.49 / 2.86 & \cellcolor{green!15}2.28 / 0.78 & \cellcolor{green!15}25.74 / 11.20\\
    \bottomrule
    \end{tabular}
      \caption{Performance of brain-to-text decoding models under different splitting methods with teacher-forcing and without teacher-forcing. The \colorbox{green!15}{green mark} and \colorbox{red!15}{red mark} denotes a method without and with text stimuli leakage correspondingly.}
  \label{teacherf}%
  \vspace{-1mm}
\end{table*}%

\begin{table*}
  \centering
    \begin{tabular}{ccccccccc}
    \toprule
    \multirow{2}{*}{\textbf{Model}} & \multirow{2}{*}{\textbf{Epoch+lr+Method}} & \multicolumn{4}{c}{\textbf{BLEU-N ($\%$)}} & \multicolumn{3}{c}{\textbf{ROUGE-1 ($\%$)}} \\
       \cmidrule(r){3-6}  \cmidrule(r){7-9}&      & \multicolumn{1}{c}{$N=1$} & \multicolumn{1}{c}{$N=2$} & \multicolumn{1}{c}{$N=3$} & \multicolumn{1}{c}{$N=4$} & \multicolumn{1}{c}{$F$} & \multicolumn{1}{c}{$P$} & \multicolumn{1}{c}{$R$} \\
    \midrule
    \multirow{5}{*}{UniCoRN}& 10+1e-3+(a) & 49.56 & 30.49 & 21.07 & 15.49 & 44.83 & 50.41 & 40.65 \\
        & 10+1e-3+(b) & 26.37 & 7.50   & 2.48  & 0.99  & 22.28 & 25.99 & 19.62 \\
        & 10+1e-3+(c) & 50.24 & 30.83 & 21.23 & 15.60  & 44.68 & 49.44 & 41.01 \\
        & 10+1e-3+(d) & 49.63 & 30.29 & 20.85 & 15.32 & 45.06 & 50.47 & 41.03 \\
        & 10+1e-3+(e) & 28.94 & 9.39  & 4.07  & 1.53  & 21.68 & 24.64 & 19.49 \\
    \midrule
    \multirow{10}{*}{UniCoRN$^*$} 
        & 20+1e-4+(a) & 50.19 & 34.25 & 25.98 & 21.00  & 46.59 & 50.36 & 43.62 \\
        & 30+1e-4+(a) & 55.46 & 40.99 & 32.85 & 27.56 & 52.08 & 55.02 & 49.68 \\
        \cmidrule(r){2-9}
        & 20+1e-4+(b) & 25.91 & 8.80   & 3.84  & 1.66  & 20.65 & 27.74 & 16.57 \\
        & 30+1e-4+(b) & 25.91 & 8.80   & 3.84  & 1.66  & 20.65 & 27.74 & 16.57 \\
        \cmidrule(r){2-9}
        & 20+1e-4+(c) & 72.44 & 60.84 & 53.35 & 47.88 & 70.52 & 74.10  & 67.53 \\
        & 30+1e-4+(c) & 72.82 & 61.42 & 53.95 & 48.44 & 71.24 & 74.41 & 68.57 \\
        \cmidrule(r){2-9}
        & 20+1e-4+(d) & 65.31 & 51.02 & 42.54 & 36.72 & 62.76 & 67.09 & 59.29 \\
        & 30+1e-4+(d) & 66.56 & 53.00  & 44.75 & 39.02 & 63.89 & 67.51 & 60.95 \\
        \cmidrule(r){2-9}
        & 20+1e-4+(e) & 32.15 & 12.34 & 5.57  & 2.45  & 24.28 & 30.43 & 20.35 \\
        & 30+1e-4+(e) & 32.15 & 12.34 & 5.57  & 2.45  & 24.28 & 30.43 & 20.35 \\
    \bottomrule
    \end{tabular}
      \caption{Generation quality of UniCoRN model for fMRI under different training settings. Here UniCoRN$^{*}$ indicates the encoder of UniCoRN is randomly initialized instead of pre-trained through signal reconstruction task.}
  \label{tab2}%
  \vspace{-3mm}
\end{table*}%

\begin{table*}
  \centering
    \begin{tabular}{ccccccccc}
    \toprule
    \multirow{2}{*}{\textbf{Model}} & \multirow{2}{*}{\textbf{Epoch+lr+Method}} & \multicolumn{4}{c}{\textbf{BLEU-N ($\%$)}} & \multicolumn{3}{c}{\textbf{ROUGE-1 ($\%$)}} \\
       \cmidrule(r){3-6}  \cmidrule(r){7-9}&      & \multicolumn{1}{c}{$N=1$} & \multicolumn{1}{c}{$N=2$} & \multicolumn{1}{c}{$N=3$} & \multicolumn{1}{c}{$N=4$} & \multicolumn{1}{c}{$F$} & \multicolumn{1}{c}{$P$} & \multicolumn{1}{c}{$R$} \\
    \midrule
    \multirow{6}{*}{UniCoRN}& 50+1e-4+(a) & 58.09 & 49.23 & 43.23 & 38.43 & 63.88 & 61.12 & 67.50 \\
        & 80+1e-4+(a) & 60.88 & 50.52   & 43.42  & 37.84  & 65.17 & 61.16 & 70.72 \\
        \cmidrule(r){2-9}
        & 50+1e-4+(c) & 52.30 & 42.89 & 36.80 & 32.17  & 57.39 & 51.09 & 67.29 \\
        & 80+1e-4+(c) & 60.78 & 55.92 & 53.18 & 51.10 & 84.64 & 63.16 & 71.50 \\
        \cmidrule(r){2-9}
        & 50+1e-4+(d) & 22.90 & 7.36  & 2.71  & 0.95  & 17.73 & 19.90 & 17.33 \\
        & 80+1e-4+(d) & 22.90 & 7.36  & 2.71  & 0.95  & 17.73 & 19.90 & 17.33 \\
    \midrule
    \multirow{6}{*}{EEG2Text} & 50+1e-4+(a) & 51.22 & 33.83 & 22.99 & 16.05  & 46.40 & 46.85 & 46.58 \\
        & 80+1e-4+(a) & 63.32 & 52.52 & 45.19 & 39.50 & 65.96 & 64.74 & 68.01 \\
        \cmidrule(r){2-9}
        & 50+1e-4+(c) & 53.83 & 38.99   & 29.57  & 23.01  & 53.64 & 54.19 & 53.56 \\
        & 80+1e-4+(c) & 65.42 &  57.56  &  52.56 &  48.60 & 73.00 & 69.99 & 77.01  \\
        \cmidrule(r){2-9}
        & 50+1e-4+(d) & 23.92 & 8.16 & 3.21 & 1.20 & 20.78 & 19.96  & 23.89 \\
        & 80+1e-4+(d) & 23.92 & 8.16 & 3.21 & 1.20 & 20.78 & 19.96  & 23.89 \\
    \bottomrule
    \end{tabular}
    \caption{Generation quality of UniCoRN and EEG2Text model for EEG under different training settings.}
  \label{tab3}%
  \vspace{-3mm}
\end{table*}%

\section{Supplementary Experiment} \label{longepoch}
\paragraph{The influence of teacher-forcing} \citet{jo2024eegtotextmodelsworking} pointed out that previous methods applied teacher-forcing during generation, which has the tendency to cause overfitting problems. Therefore, we conduct experiments on models without teacher-forcing to ensure that the affection of data leakage is not influenced by teacher-forcing. Results are shown in Table \ref{teacherf}. Without the influence of teacher-forcing decoding, splitting methods with data leakage will still lead to overestimation of model performance.

\paragraph{Experiments on longer training epochs with smaller learning rate} 
If evaluation indicators keep improving as training epochs increase, we believe part of the test set is leaked into training set and the model is overfitting. For fMRI signal, we test five current dataset splitting methods under different training settings. As shown in Table \ref{tab2}, we test two kinds of UniCoRN models. One is UniCoRN with hyper-parameters claimed in the original paper, and the other is UniCoRN$^{*}$ whose encoder is randomly initialized. Besides, UniCoRN$^{*}$ is trained with longer epochs and smaller learning rate. In method (a), (c), (d), due to text stimuli leakage, if we reduce the learning rate and extend training epochs, UniCoRN$^{*}$ performs much better than UniCoRN and its performance keeps rising with longer training epochs. As to method (b) and (e) with no text stimuli leakage, changing training epochs or learning rates makes no obvious difference to model performance. For EEG signal, the conclusion is similar as shown in Table \ref{tab3}. For method (a) and (c) with text stimuli leakage, model performance keeps rising with longer training epochs. For method (d) without text stimuli leakage, both models reach optimal performance after the first few rounds of training epochs.

\begin{algorithm*}
    \caption{Dataset splitting method for EEG signal}
    \label{al1}
    \textbf{Initialize:} Bipartite graph $\mathcal{G}_1=(\mathcal{U}, \mathcal{V}, \mathcal{E})$, $\mathcal{G}_2=(\mathcal{U}, \mathcal{V},\mathcal{E}^\prime)$ where $\mathcal{U} = \{S_i\}_{i=1}^{N}$ and $\mathcal{V}=\{T_{j}\}_{j=1}^{M}$, Adjacency matrix $A_1$ of $\mathcal{G}_1$ where $A_1[i][j] = 1$ if node $i$ and node $j$ is connected else $A_1[i][j]=0$, Adjacency matrix $A_2$ of $\mathcal{G}_2$ where $A_2[i][j] = 0$, Array $C$ where $len(C)=len(\mathcal{U})$ and $C[i]=0$\;
    \For {$u\leftarrow U_1$ \KwTo $U_N$}{
    $C_{copy} \leftarrow C$\;
    \For {$v \leftarrow A_1[u][0]$ \KwTo $A_1[u][M]$}{
    \uIf{$v =0$}{$C_{copy}[v.index] \leftarrow \infty $\;}
    }
    \emph{Minimum} $= \min (C_{copy})$\;
    $A_2[u][$\emph{Minimum}$.index] \leftarrow 1$\;
    $C[$\emph{Minimum}$.index] \leftarrow C[$\emph{Minimum}$.index] + 1$; \tcp*[f]{Make degree of nodes balanced}\
    }
    Split by subjects $\mathcal{U}$ according to default ratio\;
    $\mathcal{G}_2 = \mathcal{G}_{train} \cup \mathcal{G}_{val} \cup \mathcal{G}_{test}, \, \mathcal{U}_{train} \cap \mathcal{U}_{val} \cap \mathcal{U}_{test}  = \emptyset, \, \mathcal{V}_{train} \cap \mathcal{V}_{val} \cap \mathcal{V}_{test}  = \emptyset$\;
    \Repeat(\tcp*[f]{To three sets respectively, below is for training set}){$\mathcal{G}_{train},\mathcal{G}_{val},\mathcal{G}_{test}$ \emph{are all extended}}{
    \For {$u$ \In $\mathcal{U}$ }{
    \For {$v$ \In $\mathcal{V}$}{
    \uIf{$e=(u,v) \in \mathcal{E} $ \And $e=(u,v) \notin \mathcal{E}^\prime_{train}$ \And $u \notin \mathcal{U}_{val} \cup \mathcal{U}_{test}$}{$\mathcal{E}_{train}^{\prime} \leftarrow \mathcal{E}_{train}^{\prime} \cup \{e\}$\;}
    }
    }
    }
    \Return $\mathcal{G}_{train},\mathcal{G}_{val},\mathcal{G}_{test}$\;
\end{algorithm*}

\begin{algorithm*}
    \caption{Dataset splitting method for fMRI signal}
    \label{al2}
    \textbf{Initialize:} Bipartite graph $\mathcal{G}_1=(\mathcal{U}, \mathcal{V}, \mathcal{E})$, $\mathcal{G}_2=(\mathcal{U}, \mathcal{V},\mathcal{E}^\prime)$ where $\mathcal{U} = \{S_i\}_{i=1}^{N}$, $\mathcal{V}=\{M_{k}\}_{k=1}^{K}$, Adjacency matrix $A_1$ of $\mathcal{G}_1$ where $A_1[i][j] = 1$ if node $i$ and node $j$ is connected else $A_1[i][j]=0$, Adjacency matrix $A_2$ of $\mathcal{G}_2$ where $A_2[i][j] = 0$, Array $C$ where $len(C)=len(\mathcal{V})$ and $C[i]=0$\; 
    \For {$v\leftarrow V_1$ \KwTo $V_K$}{
    $C_{copy} \leftarrow C$\;
    \For {$u \leftarrow A_1[v][0]$ \KwTo $A_1[v][K]$}{
    \uIf{$u =0$}{$C_{copy}[u.index] \leftarrow \infty $\;}
    }
    \emph{Minimum} $= \min (C_{copy})$\;
    $A_2[v][$\emph{Minimum}$.index] \leftarrow 1$\;
    $C[$\emph{Minimum}$.index] \leftarrow C[$\emph{Minimum}$.index] + 1$; \tcp*[f]{Make degree of nodes balanced}\
    }
    Split by tasks $\mathcal{V}$ according to default ratio\;
    $\mathcal{G}_2 = \mathcal{G}_{train} \cup \mathcal{G}_{val} \cup \mathcal{G}_{test}, \, \mathcal{U}_{train} \cap \mathcal{U}_{val} \cap \mathcal{U}_{test}  = \emptyset, \, \mathcal{V}_{train} \cap \mathcal{V}_{val} \cap \mathcal{V}_{test}  = \emptyset$\;
    \Repeat(\tcp*[f]{To three sets respectively, below is for training set}){$\mathcal{G}_{train},\mathcal{G}_{val},\mathcal{G}_{test}$ \emph{are all extended}}{
    \For {$v$ \In $\mathcal{V}$ }{
    \For {$u$ \In $\mathcal{U}$}{
    \uIf{$e=(u,v) \in \mathcal{E} $ \And $e=(u,v) \notin \mathcal{E}^\prime_{train}$ \And $v \notin \mathcal{V}_{val} \cup \mathcal{V}_{test}$}{$\mathcal{E}_{train}^{\prime} \leftarrow \mathcal{E}_{train}^{\prime} \cup \{e\}$\;}
    }
    }
    }
    \Return $\mathcal{G}_{train},\mathcal{G}_{val},\mathcal{G}_{test}$\;
\end{algorithm*}

\end{document}